\documentclass[conference]{IEEEtran}
\IEEEoverridecommandlockouts
\usepackage{cite}
\usepackage{amsmath,amssymb,amsfonts}
\usepackage{algorithmic}
\usepackage{graphicx}
\usepackage{textcomp}
\usepackage{xcolor}
\usepackage{amsmath,epsfig}
\usepackage{graphicx} 
\usepackage{svg} 
\usepackage{subfigure}
\usepackage[T1]{fontenc}
\usepackage{array}
\usepackage{arydshln}
\usepackage{booktabs}
\usepackage{multirow}
\usepackage{arydshln}
\usepackage{dashrule}
\usepackage{makecell}
\usepackage{hyperref}
\usepackage[a-3u]{pdfx}

\def\BibTeX{{\rm B\kern-.05em{\sc i\kern-.025em b}\kern-.08em
    T\kern-.1667em\lower.7ex\hbox{E}\kern-.125emX}}
\begin{document}
\sloppy
\title{HQOD: Harmonious Quantization for\\
Object Detection
}
\author{\IEEEauthorblockN{
\textbf{Long Huang}\textsuperscript{1}~~
\textbf{Zhiwei Dong}\textsuperscript{1}~~
\textbf{Song-Lu Chen}\textsuperscript{1}~~
\textbf{Ruiyao Zhang}\textsuperscript{1}
\\
\textbf{Shutong Ti}\textsuperscript{1}~~
\textbf{Feng Chen}\textsuperscript{2}~~
\textbf{Xu-Cheng Yin}\textsuperscript{1\dag}
}
\IEEEauthorblockA{
\textsuperscript{1} University of Science and Technology Beijing~~
\textsuperscript{2} EEasy Technology Company Ltd.\\
\{long.huang.cn, dongz.cn\}@outlook.com~~
songluchen@ustb.edu.cn~~
ruiyaozhang@xs.ustb.edu.cn~~\\
tee.shutong@outlook.com~~
cfeng@eeasytech.com~~
xuchengyin@ustb.edu.cn
\thanks{
\textsuperscript{\dag} Corresponding author.
}
}
}

\maketitle

\begin{abstract}
Task inharmony problem commonly occurs in modern object detectors, leading to inconsistent qualities between classification and regression tasks. The predicted boxes with high classification scores but poor localization positions or low classification scores but accurate localization positions will worsen the performance of detectors after Non-Maximum Suppression. Furthermore, when object detectors collaborate with Quantization-Aware Training (QAT), we observe that the task inharmony problem will be further exacerbated, which is considered one of the main causes of the performance degradation of quantized detectors. To tackle this issue, we propose the Harmonious Quantization for Object Detection (HQOD) framework, which consists of two components. Firstly, we propose a task-correlated loss to encourage detectors to focus on improving samples with lower task harmony quality during QAT. Secondly, a harmonious Intersection over Union (IoU) loss is incorporated to balance the optimization of the regression branch across different IoU levels. The proposed HQOD can be easily integrated into different QAT algorithms and detectors. Remarkably, on the MS COCO dataset, our 4-bit ATSS with ResNet-50 backbone achieves a state-of-the-art mAP of 39.6\%, even surpassing the full-precision one. Codes
are available at \url{https://github.com/Menace-Dragon/VP-QOD}.
\end{abstract}


\section{Introduction}

\label{sec:intro}

Deep Convolutional Neural Networks (CNNs) have made remarkable strides in various applications of object detection \cite{SSD, RetinaNet, ATSS, YOLOX}. Nevertheless, these detectors with outstanding performance exhibit significant computational and parameter demands, facing challenges to efficiently run on devices with limited resources (e.g., mobile phones or drones).

In modern object detectors, the Task Inharmony (TI) problem has always been a focal point of attention \cite{TOOD, TBD, HT-CDOD}. Usually, a multi-task pipeline is used to generate both location coordinates and corresponding labels for an object, including a classification branch and a regression branch with two parallel heads. This parallel design can result in an inconsistent distribution of classification scores and regression scores (i.e., Intersection over Union, IoU). 
\begin{figure}[htbp]
\centering  
\subfigure[Baseline LSQ]{
\includegraphics[scale=0.44, trim=7.2cm 4.1cm 16.4cm 1.8cm, clip]{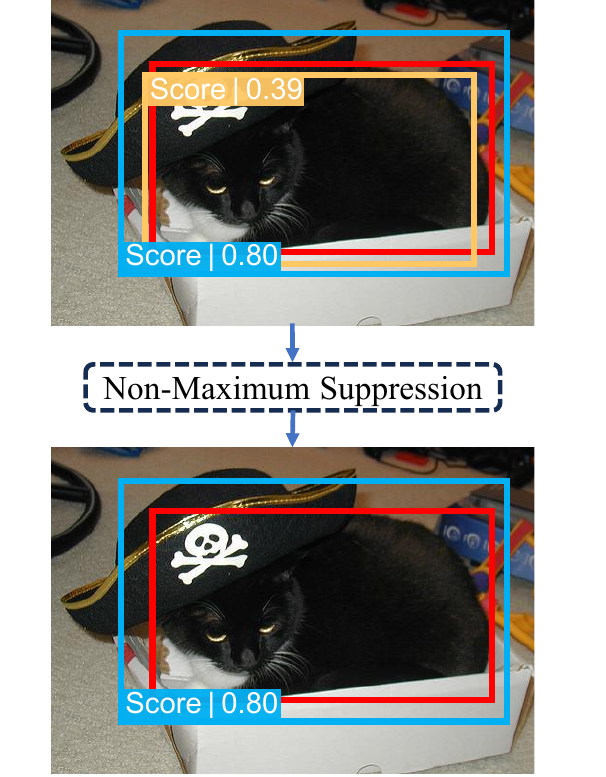}}
\subfigure[Ours]{
\includegraphics[scale=0.44, trim=18cm 4.1cm 7.2cm 1.8cm, clip]{fig/000000108244.pdf}}
\vspace{-0.4em}
\caption{The results of
object detection between the baseline LSQ \cite{LSQ} and our proposed framework. The ground truth is represented by the red bounding box, and the classification scores of the predicted boxes are explicitly indicated. Our framework enhances the harmonious relationship between the classification task and regression task, producing more accurate bboxes.
}
\label{fig:TICat}
\vspace{-0.8em}
\end{figure}
Moreover, to efficiently deploy detectors on resource-constrained devices, the application of model compression techniques is necessary. 

As one of the popular model compression techniques, Quantization-Aware Training (QAT) methods \cite{chun2023survey, DSQ, LSQ, TQT, AQD} are introduced as the widely accepted approach to achieve low-bit quantization while preserving near full-precision performance. By simulating the feedforward quantization operations during time-consuming training or fine-tuning, the network can readily adjust to the quantization noise, leading to more optimal solutions. LSQ \cite{LSQ} treats the quantization parameter, step size $s$ as a learnable parameter, enabling $s$ to adaptively learn optimal casting. TQT \cite{TQT} adopts the concept of learnable $s$ and applies it to hardware-friendly power-of-2 (PoT) scale quantization, achieving exciting results in PoT scale quantization. AQD \cite{AQD} introduces the Sync-BN in the shared detection head to alleviate the suboptimal issue of $s$ arising from the multi-level inputs. However, when quantizing low-bit object detectors, the performance still exhibits substantial gaps compared to the full-precision ones. 

In this work, we identify that the exacerbation of the TI problem under low-bit constraints is one of the primary reasons leading to the performance gap. Taking LSQ \cite{LSQ} as an example, the quantized RetinaNet \cite{RetinaNet} with ResNet-18 \cite{ResNet} backbone generates a representative instance of inconsistent bounding boxes, as shown in Fig. \ref{fig:TICat}(a). There are two inharmonious candidates (a yellow bounding box and a blue bounding box) and one ground-truth box colored in red. After the Non-Maximum Suppression (NMS) procedure, the yellow one having a high IoU but a low classification score will be suppressed by the less accurate blue one. That is to say, the suboptimal result is preserved, while the best result is overlooked. In our proposed method, this issue can be effectively alleviated by encouraging detectors to generate more harmonious samples while retaining superior detection boxes after the QAT process, as illustrated in Fig. \ref{fig:TICat}(b).

\begin{figure}[t]
\centering  
\includegraphics[scale=0.29, trim=0cm 0.2cm 0cm 0cm, clip]{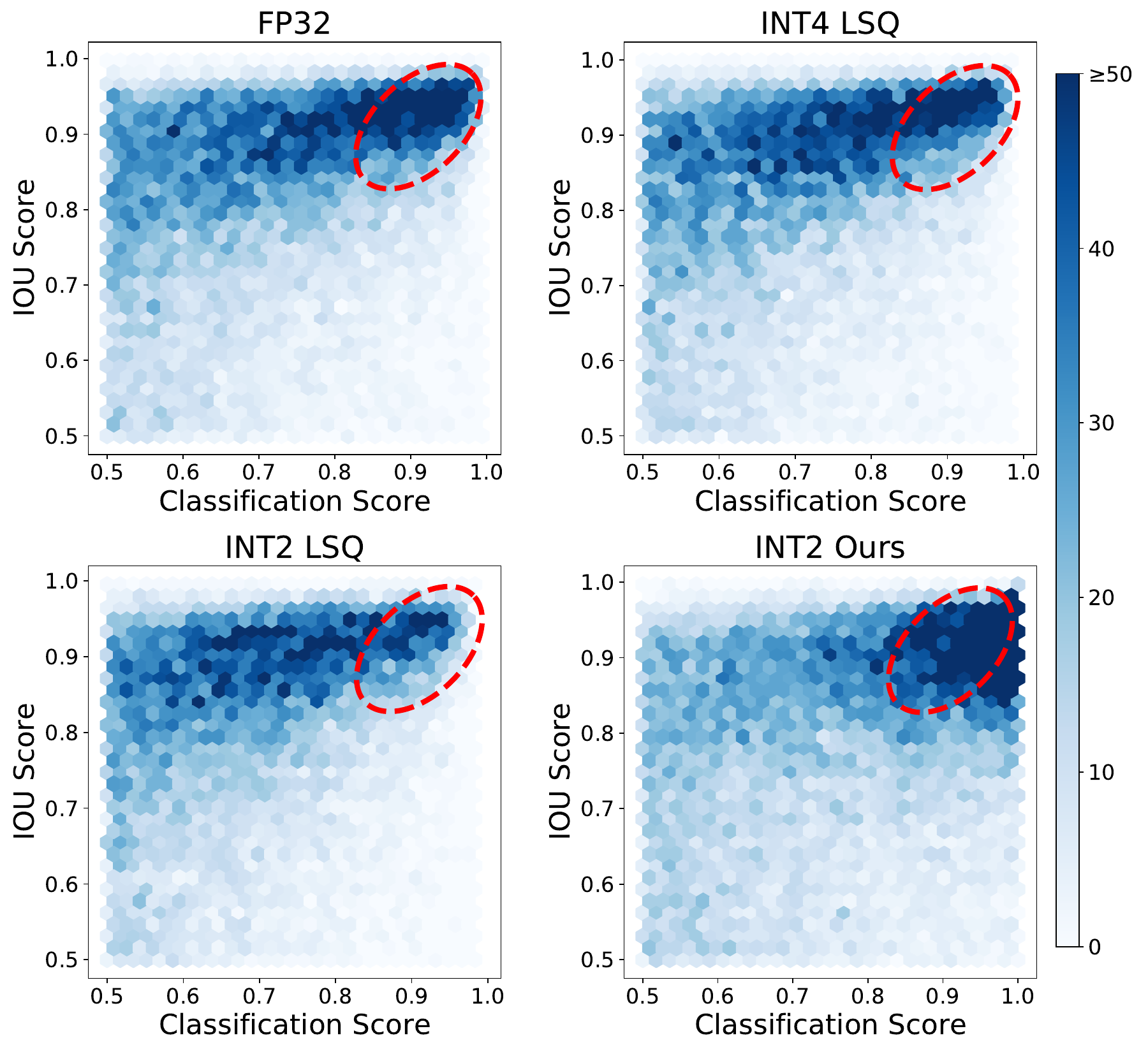}
\vspace{-0.8em}
\caption{The statistical results for true positive (TP) samples after NMS. The red dashed ellipse signifies the ideal numerical distribution, where the classification scores and IoU values are both at high levels. Note that FP32 represents a full-precision setting. INT4/2 represents that models are quantized under the 4/2-bit constraints. }
\label{fig:TiDist}
\vspace{-0.8em}
\end{figure}

In addition, we further discover that TI tends to deteriorate further when the required bit widths for QAT get lower. To illustrate this phenomenon, we yield Fig. \ref{fig:TiDist} by visualizing the statistical results of predicted true positive (TP) samples after NMS. With the decrease in bit widths, the number of high-quality prediction samples (i.e., both classification score and IoU score at a high level) exhibits notable reduction, and the statistic distributions deviate from the ideal elliptical line. This observation indicates that the relationship between the two tasks becomes increasingly inharmonious as the bit width constraints decrease.

To harmonize the optimization of tasks in QAT, we propose the Task-Correlated (TCorr) Loss to dynamically optimize the low harmony quality samples. We increase the weights of samples characterized by the wide gap between task scores while suppressing the weights of samples with a small gap between tasks. Thus, the detector is regularized toward harmonious optimization during the low-bit QAT process. Furthermore, we additionally introduce the Harmonious IoU (HIoU) loss to dynamically coordinate the contributions of samples at each IoU level, aiming to suppress the generation of low IoU samples and facilitate the generation of high IoU samples. By embedding TCorr loss and HIoU loss into QAT phase, our QAT framework, namely \textbf{H}armonic \textbf{Q}uantization for \textbf{O}bject \textbf{D}etection (HQOD), is constructed, facilitating the generation of a more harmonious and high-quality output distribution under low-bit constraints, as illustrated in Fig. \ref{fig:TiDist}. 

Extensive experiments on the PASCAL VOC \cite{VOC} dataset and MS COCO \cite{COCO} dataset demonstrate the robustness and generality of our method for remarkably improving the performance of quantized low-bit object detectors. On the PASCAL VOC benchmark, we achieve an average improvement of 0.75\% mAP for LSQ and an average improvement of 0.65\% mAP for AQD, respectively. On the MS COCO benchmark, we achieve numerous state-of-the-art results. Specifically, our 4-bit ATSS \cite{ATSS} with ResNet-50 \cite{ResNet} achieves 39.6\% mAP, outperforming the LSQ by 0.6\% mAP and surpassing the full-precision one by 0.2\% mAP. Even under extreme constraints, our 2-bit ATSS can bring an improvement of 1.4\% mAP compared to TQT. 

\vspace{0.4em}
\section{Proposed Method}
\vspace{0.2em}
\subsection{Paradigm of Quantization-Aware Training}

\label{sec:Paradigm of QAT}

To enhance comprehension of the operational logic underlying these baselines, we initially present the general paradigm of QAT. Quantizing a neural network model can be defined as a finite affine process. Given a full-precision vector $\mathbf{V} = [v_0, ..., v_{n-1}]$ and $b$ bit, the quantization function $Q_b(\cdot)$ can be formulated as follow: 
\begin{equation}
\mathbf{\hat{V}} = Q_b(\mathbf{V}) \in \{q_{0}, q_{1}, ..., q_{2^{b}-1}\}
\end{equation}
where $q_i \in R$ is the quantization level, and $2^{b}-1$ is the number of quantization levels. In this paper, a per-tensor, symmetric uniform quantization scheme is adopted for quantizing both the weight tensor and feature map (i.e., activation) tensor. Generally, QAT requires training the network with simulated quantization. In the forward propagation, the simulated quantization function can be formulated as: 
\begin{equation}
\bar{v} = clip(\lfloor \frac{v}{s} \rceil, N_{min}, N_{max}), \hat{v} = s \cdot \bar{v} 
\label{eq:qdq}
\end{equation}
where $s$ is called quantization step size, and $\lfloor \cdot \rceil$ serves as the rounding function to affine the float-point values to the nearest integers. For signed data, $N_{min} = - 2^{b-1}$ and $N_{max} = 2^{b-1}-1$. For unsigned data, $N_{min} = 0$ and $N_{max} = 2^{b} - 1$. Specifically, Eq. \eqref{eq:qdq} first quantizes values into the integer domain and then undergoes de-quantization to undo the scaling step. The effect of quantization is thus simulated while retaining the original scale of the input tensor. In the backward pass, STE \cite{STE} is generally introduced to approximate the gradients of the rounding function to 1. Therefore, the local gradient of the $\hat{v}$ with respect to $v$ can be defined as: 
\begin{equation}
\frac{\partial \hat{v}}{\partial v} = 1, N_{min} \leq \frac{v}{s} \leq N_{max}\label{eq:ste}
\end{equation}

During the QAT process, the parameters in the model can accept gradient backpropagation and updates. Thus, our subsequent proposals of task-related losses can effectively promote the updates and optimization of model parameters.
\vspace{0.2em}
\subsection{Harmonious Quantization for Object Detection}
\vspace{0.4em}
\subsubsection{Task Inharmony in Object Detection}
\ \vspace{0.4em}
\newline
\indent A standard loss of object detector can be revisited as follows:
\begin{equation}
\mathcal{L}_{\mathrm{OD}}=\frac1P(\sum_{i\in \mathrm{Pos}}^P(\mathcal{L}_{\mathrm{cls}}^i+\mathcal{L}_{\mathrm{reg}}^i)+\sum_{j\in \mathrm{Neg}}^N\mathcal{L}_{\mathrm{cls}}^j) \label{eq:standard loss}
\end{equation}
where $P$ and $N$ are the number of positive and negative samples, respectively. $\mathcal{L}_{\mathrm{cls}}$ and $\mathcal{L}_{\mathrm{reg}}$ denote the optimization loss for classification and bounding box regression, depending on different models, respectively.

The classification and regression branches are trained with separate objective functions for each positive sample, resulting in significant inconsistency between the classification and regression tasks. In other words, $\mathcal{L}_{\mathrm{cls}}$ encourages the model to learn high classification scores during training, irrespective of the localization scores, while $\mathcal{L}_{\mathrm{reg}}$ aims to improve localization ability, regardless of classification results. Consequently, the predicted classification scores and localization scores become detached from each other, leading to inconsistent detections. These inconsistent detections may exhibit high classification scores but low IoUs, or low classification scores but high IoUs, ultimately compromising the overall detection performance after NMS.

\vspace{0.8em}
\subsubsection{Task-Correlated Loss}
\begin{figure}[t]
\centering  
\subfigure[$\beta_{\mathrm{cls}}=0.5  \hspace{0.5em},\hspace{0.5em} \beta_{\mathrm{reg}}=0.5$]{
\includegraphics[width=4.2cm,height=3.1cm]{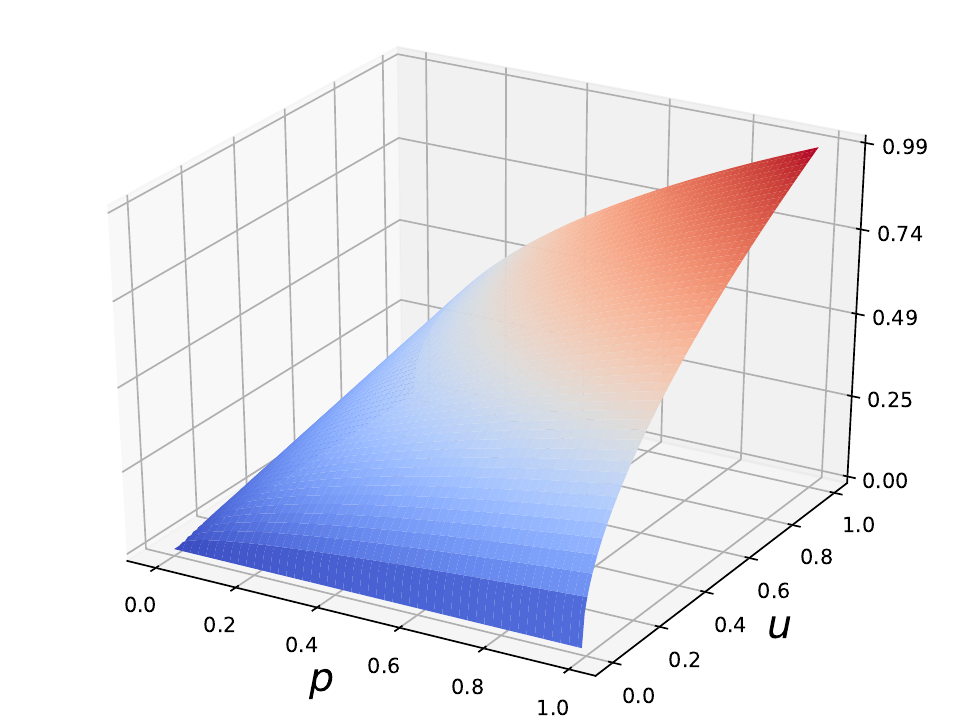}}
\subfigure[$\beta_{\mathrm{cls}}=u  \hspace{0.5em},\hspace{0.5em} \beta_{\mathrm{reg}}=p$]{
\includegraphics[width=4.2cm,height=3.1cm]{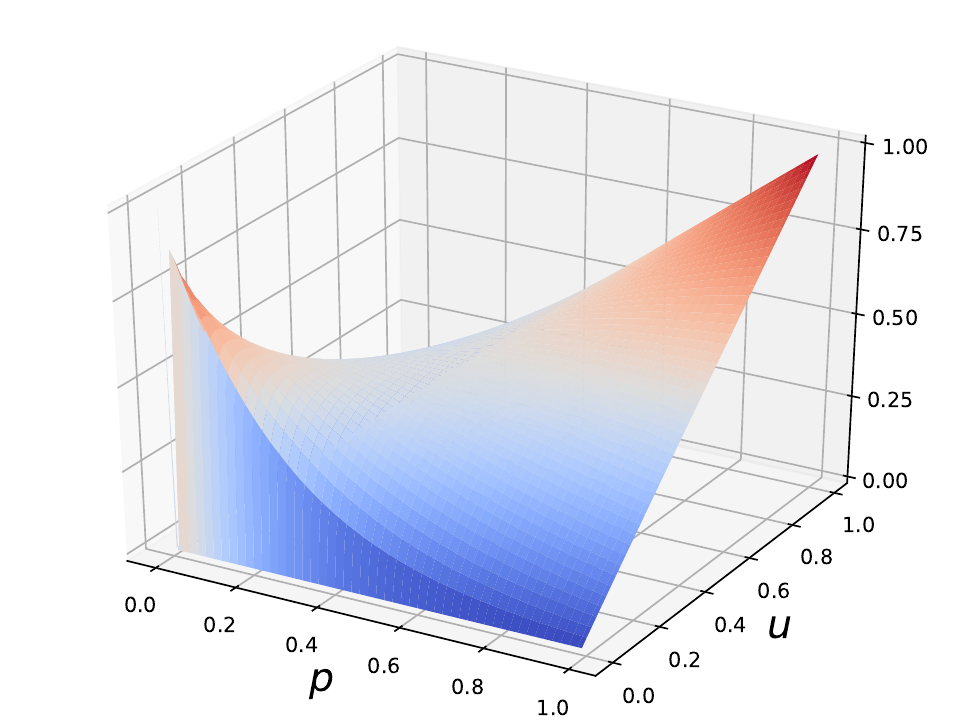}}
\vspace{-0.4em}
\caption{Visualization of the task correlation indicator $c$ with different setting of $\beta_{\mathrm{cls}}$ and $ \beta_{\mathrm{reg}}$ in Eq. \eqref{eq:task correlation indicator}.}
\label{fig:different factors}
\vspace{-0.2em}
\end{figure}
\ \vspace{0.4em}
\newline
\indent During the QAT procedure, the standard detection loss in Eq. \eqref{eq:standard loss} will be enabled to update and optimize weights and quantization step size. However, noise introduced by quantization will obstruct the process of optimization, further exacerbating the task inharmony problem, especially in low-bit quantization. To tackle this problem, we first introduce a task correlation indicator to properly elucidate the harmony quality of a sample, defined as follows:
\begin{equation}
\label{eq:task correlation indicator}
c_{i}=p_{i}^{\beta_{\mathrm{cls}}}*u_{i}^{\beta_{\mathrm{reg}}}
\end{equation}
where $p_{i}$ denotes the classification score of $i$-th positive sample, output by classification branch. $u_{i}$ is the IoU score of $i$-th positive sample predicted from the regression branch. Note that the range of both classification score and IoU is $[0, 1]$, therefore the range of the task correlation indicator is $[0, 1]$. This indicator has two basic characteristics. Firstly, it takes into consideration the score quality of the two branches. When $c=1$, both $p$ and $u$ are equal to 1, indicating that both the classification and regression branches produce high-quality samples. When $c=0$, it signifies that at least one of the scores of branches is 0, implying that the model outputs low-quality samples. Secondly, it also characterizes the consistency between the two branches. A smaller gap in scores between the two branches produces a higher $c$, while a larger difference in the output results of the two branches produces a lower $c$. $\beta_{\mathrm{cls}}$ and $\beta_{\mathrm{reg}}$ are two dynamic factors rather than two constants, and we define them as:
\begin{equation}
\label{eq:dynamic factors}
\beta_{\mathrm{cls}}=u_{i} \hspace{0.5em},\hspace{0.5em} \beta_{\mathrm{reg}}=p_{i}
\end{equation}

By substituting Eq. \eqref{eq:dynamic factors} into Eq. \eqref{eq:task correlation indicator}, we can obtain the functional effect as illustrated in Fig. \ref{fig:different factors}. When both $p$ and $u$ are low, $c$ does not exhibit a low value under the influence of dynamic factors; instead, $c$ tends to increase. We posit that when both task scores of a sample are low, there is less need to excessively discuss the harmony issue of that sample, while $c$ will attain a relatively higher level. In this manner, only two types of samples will be given special attention and assigned lower $c$, preventing a large number of low-quality positive samples from dominating the harmonious optimization process. One type consists of the sample with a high score on one of the tasks but with a significant score gap between tasks. The other type includes the sample that exhibits a small score gap between tasks but with both task scores hovering around 0.5. Subsequently, we propose Task-Correlated (TCorr) loss to focus on and optimize the samples with lower $c$, as shown in Eq. \eqref{eq:TCorr}.
\begin{equation}
\label{eq:TCorr}
\mathcal{L}_{\mathrm{TCorr}}^i=\alpha(e^{-c_i}-e^{-1}) \hspace{0.6em},\hspace{0.6em} \alpha=(1+|p_i-u_i|)
\end{equation}
where $\alpha$ serves as the reweighting parameter and does not accept gradient backpropagation. The magnitude of the score gap between tasks directly corresponds to the increasing emphasis on the loss function.
\vspace{0.8em}
\subsubsection{Harmonious IoU Loss}
\begin{figure}[t]
\centering  
\includegraphics[scale=0.18]{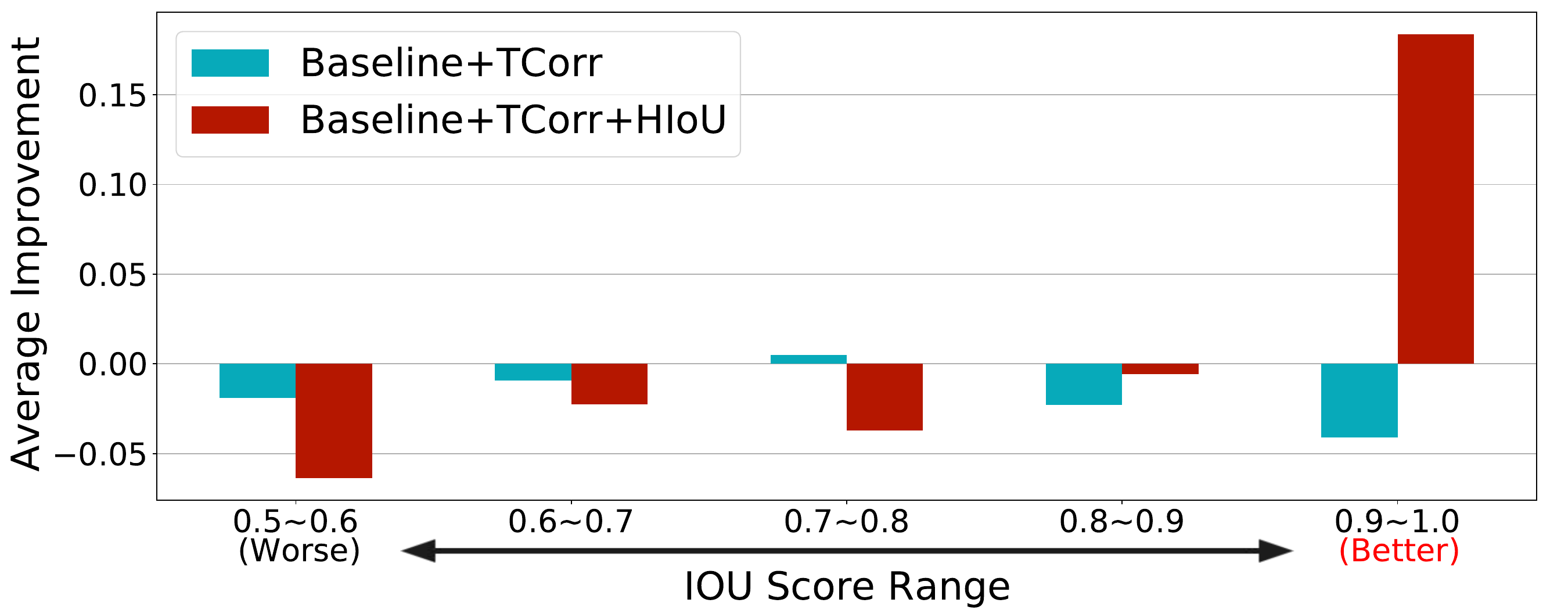}
\vspace{-0.4em}
\caption{The average improvement compared to baseline LSQ \cite{LSQ} for positive samples from different IoU intervals.}
\label{fig:IoUVariation}
\vspace{0.2em}
\end{figure}

\ \vspace{0.4em}
\newline
\indent After using TCorr, there is a reduction in the proportion of positive samples at lower IoU levels, aligning with our expected outcome as illustrated in Fig. \ref{fig:IoUVariation}. However, we observe a slight decrease in the proportion of positive samples at higher IoU levels, particularly within the IoU range of 0.9 to 1.0, contrary to our expectations. We consider that this phenomenon arises due to the dominance of samples with lower IoU levels in the optimization process of the regression branch, thereby dominating the directions of gradient updates. Generally, detectors tend to generate a substantial number of positive bounding boxes for the optimization of localization losses. However, a significant proportion of these samples exhibit low IoU levels, consequently exerting a predominant influence on the optimization direction of the localization loss. To mitigate this biased optimization, Harmonious IoU (HIoU) loss is introduced and defined as follows:
\begin{equation}
\label{eq:HIoU}
\mathcal{L}_{HIoU}^{i}=(1+u_{i})^{\gamma}(1-u_{i})
\end{equation}

In our experiment, the hyperparameter $\gamma$ is set to 0.8. The adaptive weight $(1+u_{i})^{\gamma}$ is utilized to harmonize the localization loss across various IoU levels. Thus, HIoU loss enhances the weighting of the localization loss for samples with high IoU and simultaneously suppresses the weighting of samples with low IoU through a dynamic scaling factor. By additionally utilizing HIoU loss, the model exhibits a reduction in the output of samples with low IoU quality while generating more samples with higher IoU quality, as depicted in Fig. \ref{fig:IoUVariation}.
\vspace{0.8em}
\subsubsection{Overall Optimization Objective}
\ \vspace{0.4em}
\newline
\indent By incorporating the TCorr loss and the HIoU loss with Eq. \eqref{eq:standard loss}, the overall optimization objective of the proposed \textbf{H}armonious \textbf{Q}uantization for \textbf{O}bject \textbf{D}etection framework can be written as follows:
\begin{equation}
\mathcal{L}_{\mathrm{HQOD}}=\mathcal{L}_{\mathrm{OD}}
+\frac1P\sum_{i\in \mathrm{Pos}}^P(\mathcal{L}_{\mathrm{Tcorr}}^i+\sigma\mathcal{L}_{\mathrm{HIoU}}^i)
\label{eq:hqod}
\end{equation}

The trade-off parameter $\sigma$ is set as 1.5 in our experiments. Detectors are trained by optimizing all these losses in an end-to-end manner during the QAT procedure. 

\vspace{0.4em}
\section{Experiments}
\vspace{0.2em}
\subsection{Experimental Settings}
\textbf{Datasets.} We conduct extensive experiments on two widely used object detection datasets, MS COCO \cite{COCO} and PASCAL VOC \cite{VOC}. For MS COCO, models are trained on the 118k training split and evaluated on the minimal split. For PASCAL VOC, the union of VOC2007 trainval and VOC2012 trainval is utilized for training, while the evaluation is performed on the VOC2007 test split. 
Both two datasets are evaluated by standard MS COCO metrics with mean average precision (mAP), $\mathrm{AP}_{50}$, and $\mathrm{AP}_{75}$
, which displays more comprehensive performance details. 

\textbf{Object Detection Baselines.}
Four popular object detectors are used in our experiments, including RetinaNet \cite{RetinaNet}, ATSS \cite{ATSS}, and two lightweight models, SSD-lite \cite{SSD} and YOLOX-tiny \cite{YOLOX}. ResNet-18 \cite{ResNet}, ResNet-50 \cite{ResNet}, MobileNetV2 \cite{MobileNetv2}, and CSPDarknet \cite{CSPNet}, pre-trained on the standard classification task \cite{Imagenet}, are used as our backbone networks. All full-precision models on MS COCO and training process are obtained from open-source code, MMDetection \cite{MMDetection}.

\textbf{Quantization Baselines.} 
LSQ \cite{LSQ}, TQT \cite{TQT}, and AQD \cite{AQD} are followed as QAT baseline methods. LSQ and TQT are implemented by MQBench \cite{MQBench}.
For a fair comparison, we implement AQD \cite{AQD} based on MQBench,
eliminating the influences from different quantization settings. Note that we only implement the essence of AQD, which introduces Sync-BN \cite{AQD} in the head of the Detector.

\textbf{QAT Settings.} We adopt per-tensor uniform symmetric quantization as mentioned in Section \ref{sec:Paradigm of QAT} and quantize all the convolutional layers of a model. When a convolutional layer is quantized, it means that both the input feature maps and the layer weights are quantized under the same bit constraint. The first and last layers of all the detectors are only quantized to 8-bit widths. 
For the workflow and hyper-parameters of QAT, please refer to Section A of the supplementary materials.


\begin{table}[t]
    \small       
    
    \caption{Performance for RetinaNet with backbone ResNet-18 on PASCAL VOC. * represents our re-implementation. BW represents the global bit width.}         
    \label{tab:LSQ-VOC}         
    \centering      
    \setlength{\tabcolsep}{3.2mm}{\begin{tabular}{cccccc}
        \toprule
        Method & BW  & mAP & $\mathrm{AP}_{50}$ & $\mathrm{AP}_{75}$  \\
        \hline
        \midrule
        Full-precision   & FP32  & 50.4 & 78.9 & 53.8  \\
                \cdashline{1-5}[4pt/4pt]
                \\[-8pt]
                    LSQ\cite{LSQ} & INT4  & 50.2 & 78.6 & 53.4   \\
                    LSQ+HQOD & INT4  & \textbf{50.8} & {78.7} & {53.9}   \\
                \cdashline{1-5}[4pt/4pt]
                \\[-8pt]
                    LSQ\cite{LSQ}  & INT2  & 47.9 & {75.3} & 50.6   \\
                    LSQ+HQOD  & INT2  & \textbf{48.8} & 75.3 & {51.5}   \\
        \midrule
        Sync-BN \cite{AQD}   & FP32  &51.3&80.1&55.1\\
                \cdashline{1-5}[4pt/4pt]
                \\[-8pt]
                    AQD*\cite{AQD} & INT4  & 50.7&79.5&54.3   \\
                    AQD*+HQOD & INT4  & \textbf{51.3}&79.5&55.6   \\
                \cdashline{1-5}[4pt/4pt]
                \\[-8pt]
                    AQD*\cite{AQD}  & INT2  & 48.1&75.6&51.0   \\
                    AQD*+HQOD  & INT2  & \textbf{48.8}&75.2&52.8   \\
        \bottomrule
    \end{tabular}}
    \vspace{-0.2em}
    
\end{table}

\vspace{0.2em}
\subsection{Comparisons to SOTA Quantization Methods}
\label{sec:COCOExp}
The quantization performance of 4/2-bits detectors is shown in Tables  \ref{tab:LSQ-VOC}, \ref{tab:4-bit-COCO}, and \ref{tab:2-bit-COCO}.
Additionally, we include performance results of full-precision detectors for comparison.
\vspace{0.8em}
\subsubsection{Experiments on PASCAL VOC}
\ \vspace{0.4em}
\newline
\indent From Table \ref{tab:LSQ-VOC}, we can observe that our HQOD demonstrates a significant improvement in mAP across different bit widths. Specifically, under the 4-bit configuration, our HQOD can outperform both LSQ and AQD by 0.6\% mAP. Even under the extreme 2-bit constraints, a relative enhancement of 0.9\% and 0.7\% 
 mAP is obtained compared to the baseline LSQ and AQD, respectively. 

\begin{table}[t]
    \small       
    \caption{Performance for 4-bit detectors on MS COCO.}         
    \label{tab:4-bit-COCO}         
    \centering      
    \setlength{\tabcolsep}{1.26mm}{\begin{tabular}{cccccc}
        \toprule
        Model         & Method & BW  & mAP & $\mathrm{AP}_{50}$ & $\mathrm{AP}_{75}$  \\
        \hline
        \midrule
        \multirow{6}{*}{\makecell[c]{RetinaNet\\(ResNet-18)}} &Full-precision   & FP32 &31.7	&49.6&33.4   \\
        \cdashline{2-6}[4pt/4pt]
                \\[-8pt]
                & LSQ \cite{LSQ} & INT4&31.4&49.3&33.0  \\
                & LSQ+HQOD & INT4&\textbf{32.5}&{50.0}&{34.2}    \\
        \cline{2-6}
                \\[-8pt]
         &Sync-BN \cite{AQD}   & FP32 &32.0&49.7&33.8   \\
        \cdashline{2-6}[4pt/4pt]
                \\[-8pt]
                & AQD* \cite{AQD} & INT4&32.1&50.4&33.9  \\
                & AQD*+HQOD & INT4&\textbf{33.1}&50.9&35.0    \\
        \midrule
        \multirow{3}{*}{\makecell[c]{RetinaNet\\(ResNet-50)}} &Full-precision   & FP32&37.4&56.7&39.6   \\
        \cdashline{2-6}[4pt/4pt]
                \\[-8pt]
                    & LSQ\cite{LSQ} & INT4&35.1&53.9&37.3  \\
                    & LSQ+HQOD & INT4&\textbf{36.9}&{55.4}&{39.5}    \\
        \midrule
        \multirow{5}{*}{\makecell[c]{ATSS\\(ResNet-50)}} &Full-precision   & FP32&39.4&57.6&42.8   \\
        \cdashline{2-6}[4pt/4pt]
                \\[-8pt]
                & LSQ\cite{LSQ} & INT4&39.0&57.8&41.9  \\
                & LSQ+HQOD & INT4&\textbf{39.6}&{58.1}&{42.5}    \\
        \cdashline{2-6}[4pt/4pt]
                \\[-8pt]
                & TQT\cite{TQT} & INT4  &39.0&57.9&41.9  \\
                & TQT+HQOD & INT4  &\textbf{39.5}&{58.0}&{42.4}    \\
        \midrule
        \multirow{3}{*}{\makecell[c]{SSD-lite\\(MobileNetV2)}} &Full-precision   & FP32  &21.3&35.4&21.8   \\
        \cdashline{2-6}[4pt/4pt]
                \\[-8pt]
                & LSQ\cite{LSQ} & INT4  &18.6&31.6&19.0  \\
                & LSQ+HQOD & INT4  &\textbf{18.8}&{31.8}&{19.0}    \\
                
        \midrule
        \multirow{3}{*}{\makecell[c]{YOLOX-tiny\\(CSPDarknet)}} &Full-precision   & FP32  &31.8&49.1&33.8   \\
        \cdashline{2-6}[4pt/4pt]
                \\[-8pt]
                    & LSQ\cite{LSQ} & INT4  &24.2&40.2&25.5  \\
                    & LSQ+HQOD & INT4  &\textbf{24.7}&{41.2}&{25.8}    \\
        \bottomrule
    \end{tabular}}
    \vspace{-0.8em}
\end{table}

\vspace{0.8em}
\subsubsection{Experiments on MS COCO}
\ \vspace{0.4em}
\newline
\indent Results for 4-bit detectors are presented in Table \ref{tab:4-bit-COCO}. For RetinaNet with ResNet-18 backbone, our HQOD loss can improve the mAP by 1.1\%, surpassing the performance of FP32 by 0.8\%. Based on AQD, our method can even achieve the best 33.1\% mAP, surpassing the original full-precision version by an additional 1.4\%. For RetinaNet with deeper backbone ResNet-50, our method also outperforms LSQ by 1.8\%. On ATSS with ResNet-50, HQOD loss can improve the mAP of LSQ and TQT by 0.6\% and 0.5\%, respectively. Especially when the IoU threshold is 0.75 ($\mathrm{AP}_{75}$), the improvement is also significant, gaining 0.6\% and 0.5\%, respectively. These results prove that the QAT methods trained with our HQOD loss can produce more accurate bboxes. In the lightweight detectors with constrained learning capabilities, our proposed method yields respective performance improvements of 0.2\% and 0.5\% for SSD-lite and YOLOX-tiny, as compared to LSQ.
\begin{table}[t]
    \small       
    \vspace{0.8em}
    
    \caption{Performance for 2-bit detectors on MS COCO.
}         
    \label{tab:2-bit-COCO}         
    \centering      
    \setlength{\tabcolsep}{1.26mm}{\begin{tabular}{cccccc}
        \toprule
        Model         & Method & BW  & mAP & $\mathrm{AP}_{50}$ & $\mathrm{AP}_{75}$  \\
        \hline
        \midrule
        \multirow{7}{*}{\makecell[c]{RetinaNet\\(ResNet-18)}} &Full-precision   & FP32 &31.7	&49.6&33.4   \\
        \cdashline{2-6}[4pt/4pt]
                \\[-8pt]
                & LSQ\cite{LSQ} & INT2 &29.3&46.7&30.3  \\
                & LSQ+HarDet* \cite{HarDet} & INT2 &28.8&43.1&30.8  \\
                & LSQ+HQOD & INT2 &{\bf30.7}&47.4&32.3    \\
        \cline{2-6}
                \\[-8pt]
        & Sync-BN \cite{AQD} & FP32  &32.0&49.7&33.8   \\
                \cdashline{2-6}[4pt/4pt]
                \\[-8pt]
                 & AQD*\cite{AQD} & INT2  &28.9&46.2&30.4  \\
                 & AQD*+HQOD & INT2  &\textbf{30.1}&46.7&31.8   \\
        \midrule
        \multirow{3}{*}{\makecell[c]{ATSS\\(ResNet-50)}} &Full-precision   & FP32  &39.4&57.6&42.8   \\
                \cdashline{2-6}[4pt/4pt]
                \\[-8pt]
                & TQT\cite{TQT} & INT2  &33.5&51.1&35.4  \\
                & TQT+HQOD & INT2  &\bf34.8	&51.9&37.1    \\
        \bottomrule
    \end{tabular}}
    \vspace{-0.2em}
    
\end{table}
We extend the experiments to a more challenging 2-bit constraint. As shown in Table \ref{tab:2-bit-COCO}, both LSQ and AQD are evaluated on the RetinaNet framework, and the proposed HQOD loss can achieve an additional gain of 1.4\% and 1.2\% mAP, respectively. We also re-implement the HarDet \cite{HarDet} method as a comparison, which aims to promote harmony between tasks for full-precision detectors. However, HarDet is significantly poor in 2-bit constraint, with a 0.5\% degradation in mAP. Then for TQT on ATSS, our HQOD loss can further improve the mAP performance by 1.3\%, with 0.8\% and 1.7\% recovery in $\mathrm{AP}_{50}$ and $\mathrm{AP}_{75}$, respectively. In summary, our proposed method has the potential to further boost the performance of state-of-the-art QAT methods.

    
    

\begin{table}[t]
    
    \caption{Ablation Study on RetinaNet with backbone ResNet-18 under 2-bit constraint.}         
    \label{tab:Ablation}         
    \centering      
    \setlength{\tabcolsep}{4mm}{\begin{tabular}{lcccc}
        \toprule
         Method   & mAP & $\mathrm{AP}_{50}$ & $\mathrm{AP}_{75}$  \\
        \hline
        \midrule
         LSQ      &29.3&46.7&30.3    \\
        \hdashline
                \\[-8pt]
         +TCorr         &29.9&47.6&31.6   \\
         +TCorr+HIoU   &{\bf30.7}&47.4&32.3  \\
        \bottomrule
    \end{tabular}}
    \vspace{1em}
    
\end{table}


\vspace{0.2em}
\subsection{Ablation Study}

We utilize LSQ as the baseline and employ a 2-bit RetinaNet with ResNet-18 detector to conduct ablation experiments on the MS COCO dataset. As shown in Table \ref{tab:Ablation}, our proposed TCorr loss improves mAP by 0.6\%. After additionally introducing Harmonious IoU (HIoU) loss, the extended improvement of 0.8\% is achieved, resulting in a final mAP of 30.7\%. 

\begin{figure}[t]
\centering  
\includegraphics[scale=0.18]{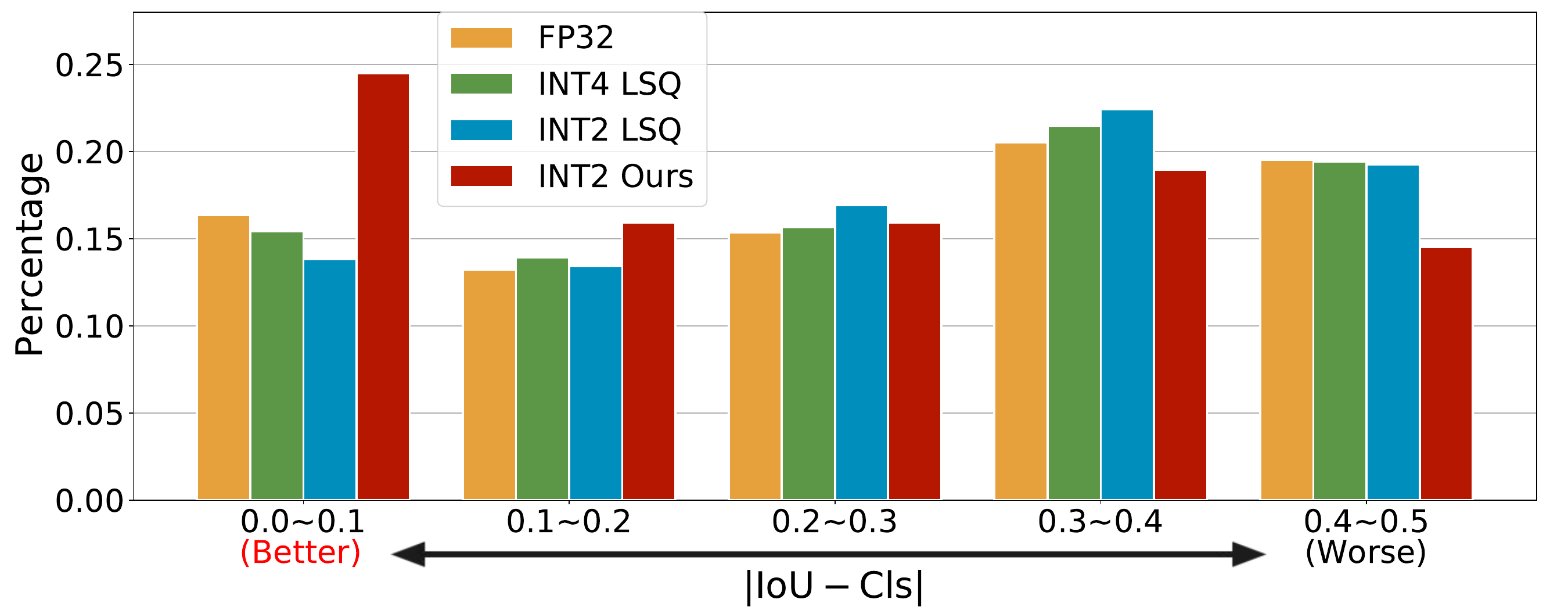}
\vspace{-0.4em}
\caption{The distribution of the gap value between IoU and classification score (Cls) based on RetinaNet with ResNet-18.}
\label{fig:har_quality}
\end{figure}
\vspace{0.2em}
\subsection{Quantitative Analysis}
\begin{figure}[t]
\centering  
\subfigure{
\begin{minipage}[b]{0.14\linewidth}
\scriptsize
\hfill w/o HQOD 

\rotatebox{90}{~~~~~~~~~~~~~~~~~}

\hfill w/ HQOD 

\rotatebox{90}{~~~~~~~~~~~~~~~~~}

\hfill w/o HQOD 

\rotatebox{90}{~~~~~~~~~~~~~~~~~}

\hfill w/ HQOD 

\rotatebox{90}{~~~~~~~~~}
\end{minipage}
}
\hspace{-10pt}
\subfigure[LSQ]{
\begin{minipage}[b]{0.27\linewidth}
\includegraphics[width=1\linewidth]{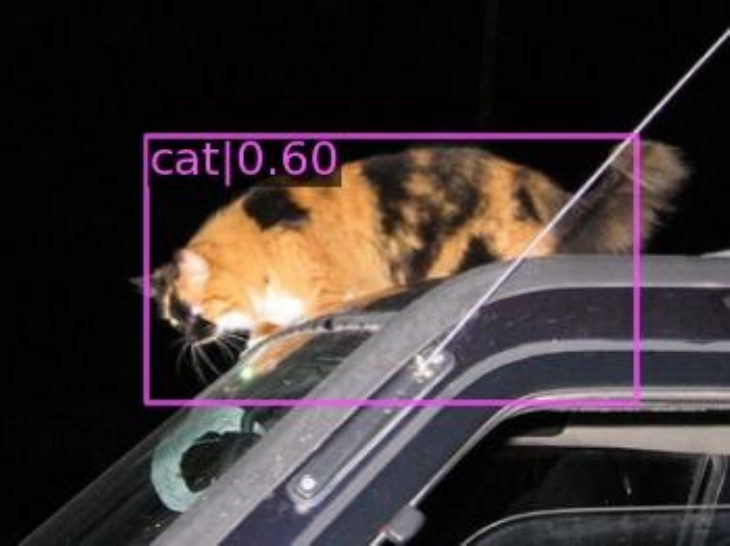}\vspace{1pt}
\includegraphics[width=1\linewidth]{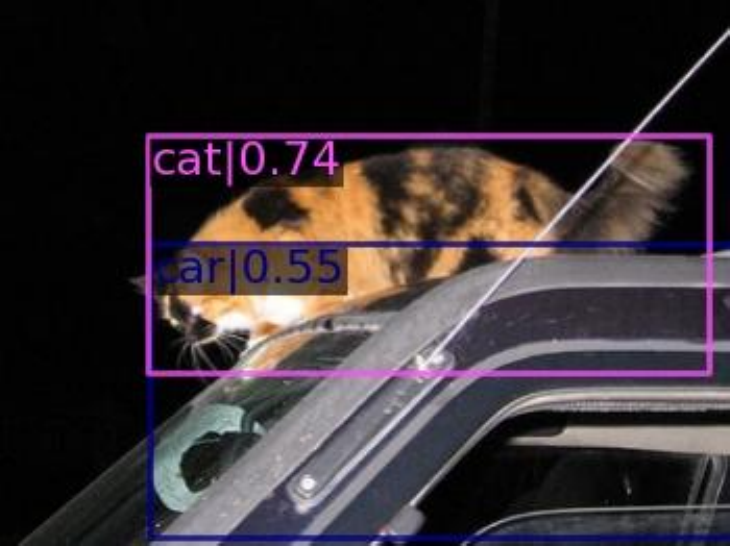}\vspace{1pt}
\includegraphics[width=1\linewidth]{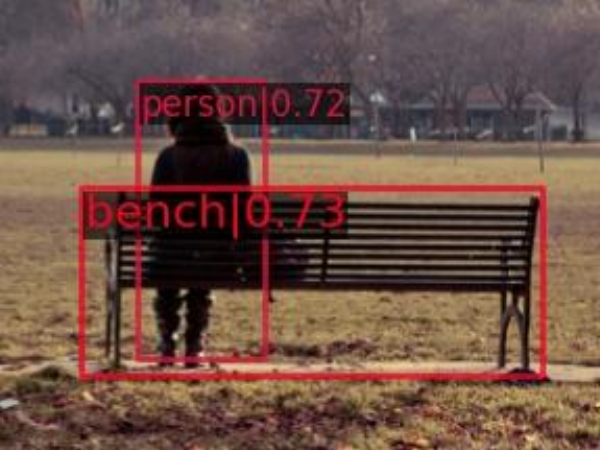}\vspace{1pt}
\includegraphics[width=1\linewidth]{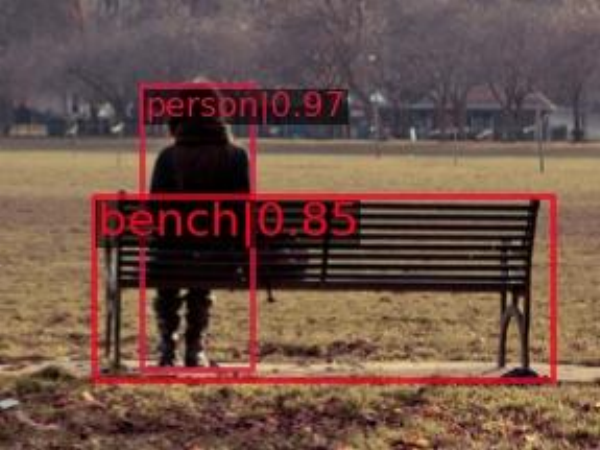}\vspace{1pt}
\end{minipage}
}
\hspace{-12pt}
\subfigure[AQD]{
\begin{minipage}[b]{0.27\linewidth}
\includegraphics[width=1\linewidth]{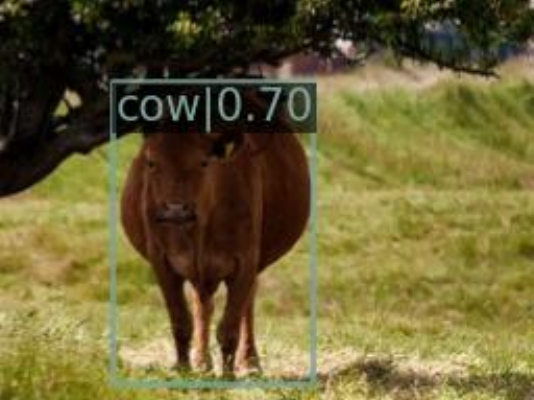}\vspace{1pt}
\includegraphics[width=1\linewidth]{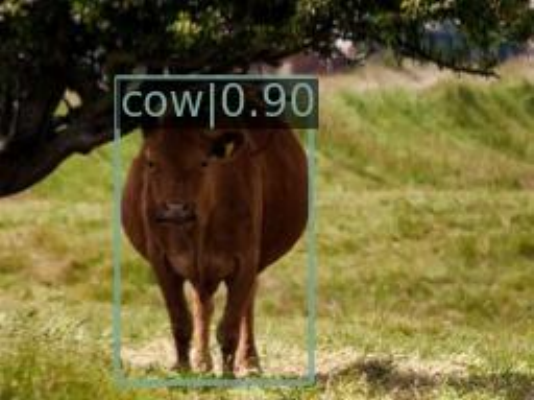}\vspace{1pt}
\includegraphics[width=1\linewidth]{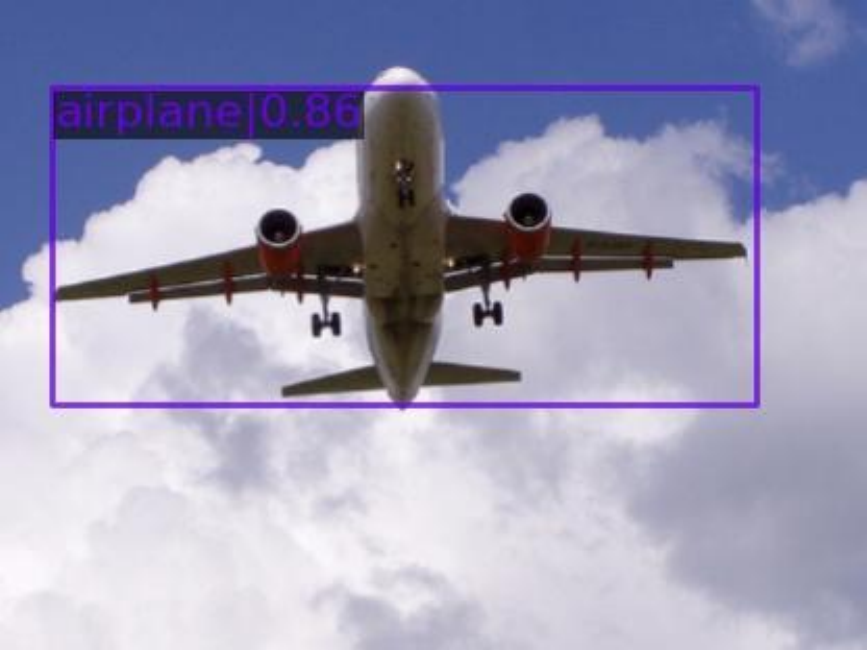}\vspace{1pt}
\includegraphics[width=1\linewidth]{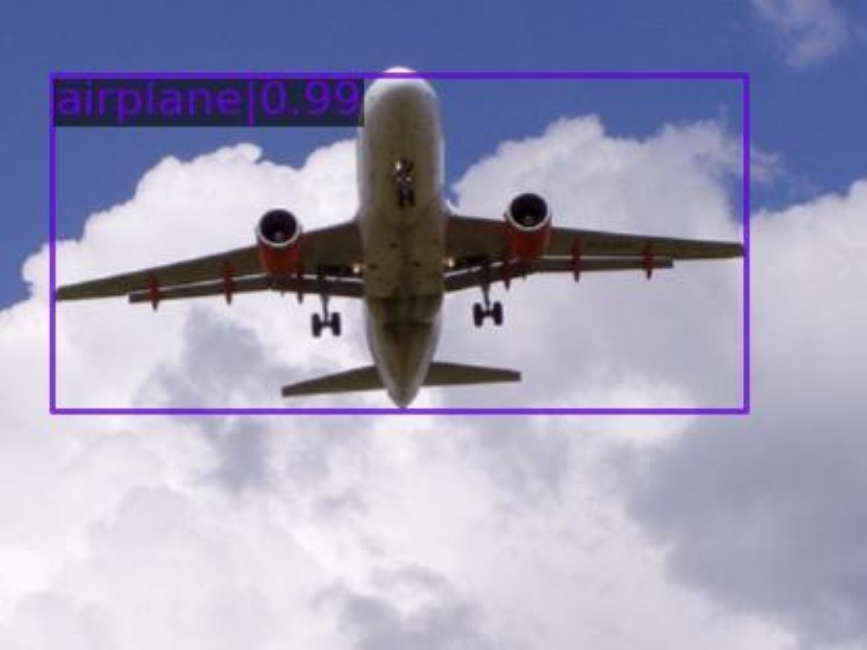}\vspace{1pt}
\end{minipage}
}
\hspace{-12pt}
\subfigure[TQT]{
\begin{minipage}[b]{0.27\linewidth}
\includegraphics[width=1\linewidth]{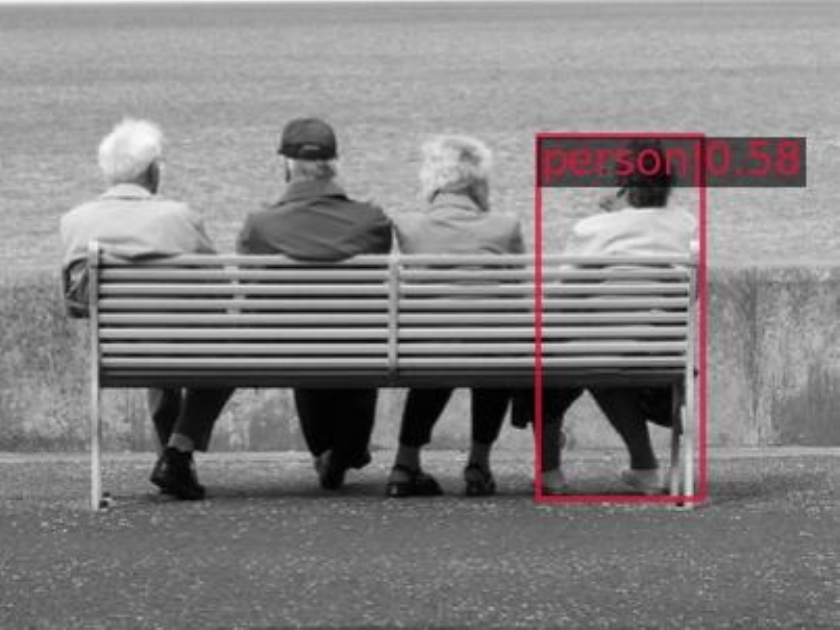}\vspace{1pt}
\includegraphics[width=1\linewidth]{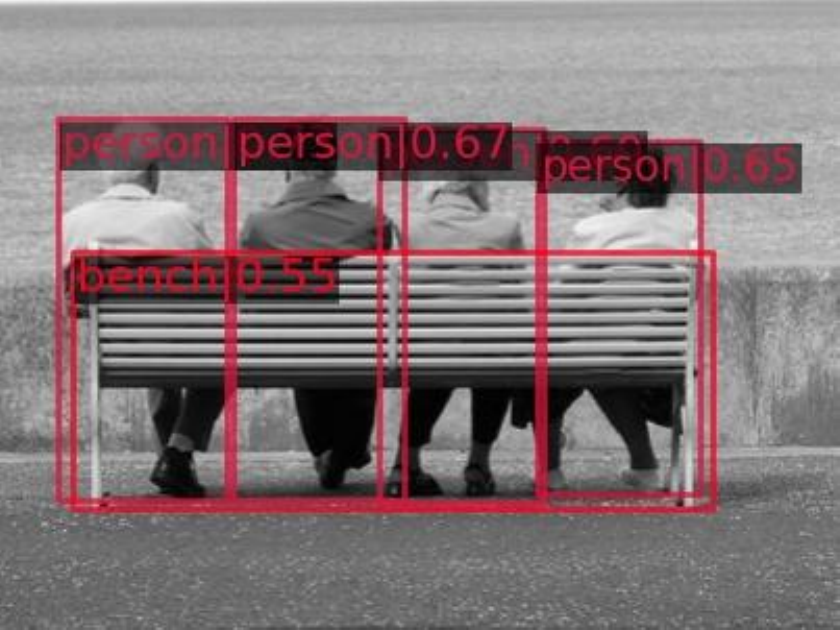}\vspace{1pt}
\includegraphics[width=1\linewidth]{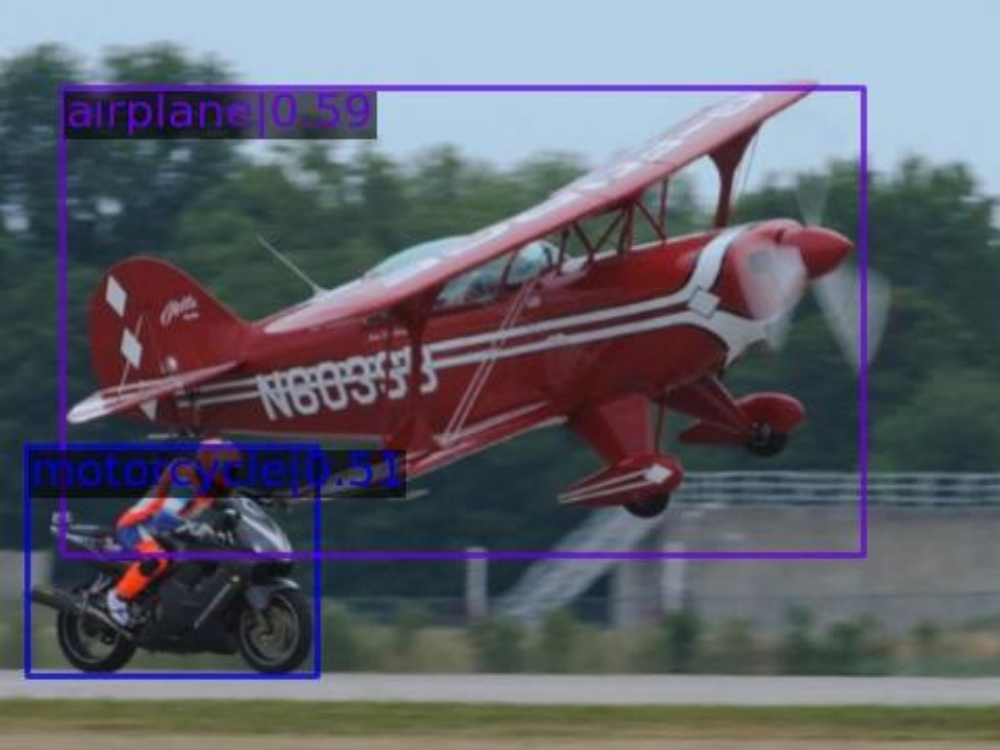}\vspace{1pt}
\includegraphics[width=1\linewidth]{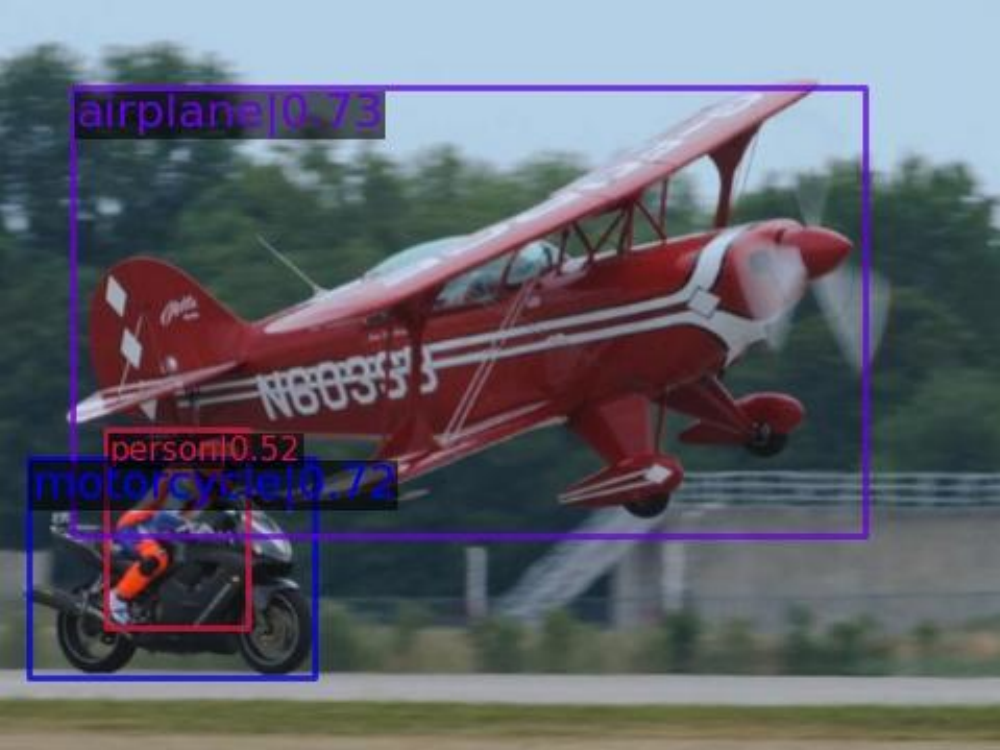}\vspace{1pt}
\end{minipage}
}

\vspace{-0.4em}
\caption{Qualitative comparisons between the state-of-the-art QAT methods with or without our HQOD framework. Our HQOD framework concurrently enhances the accuracy of both classification and regression results, producing more harmonious samples.
}
\label{fig:predictions}
\vspace{-0.8em}

\end{figure}

We conduct quantitative analysis on the MS COCO dataset. Utilizing the absolute gap between the classification score and IoU as the evaluation metric, the statistics of TP samples can be depicted in Fig. \ref{fig:har_quality}. 
After LSQ, the proportion of samples with score gaps from 0 to 0.1 significantly decreases, while the proportion of gaps ranging from 0.2 to 0.4 increases. The degradation becomes more pronounced with the decrease in bit width. These observations indicate a further exacerbation of task inharmony after low-bit QAT. After introducing our HQOD method, the proportion of gaps from 0 to 0.1 significantly increases, and inharmonious samples are effectively dampened. We also provide qualitative comparisons between state-of-the-art QAT algorithms and our method under 2-bit constraints, as illustrated in Fig. \ref{fig:predictions}. Promoting the optimization of detectors toward task harmony during the QAT phase is imperative. Our HQOD framework enables low-bit detectors to generate more accurate detection bounding boxes, effectively alleviating the inharmony between classification scores and localization scores. For more analysis, please refer to Section B of the supplementary materials.

\vspace{0.4em}
\section{Conclusion}
In this work, we identify that the task inharmony problem is exacerbated when the object detectors undergo quantization-aware training (QAT) and more pronounced with the bit width decrease, which is one of the primary issues leading to the performance degradation of the quantized detectors. To foster a more harmonious QAT process, we propose the Harmonious Quantization for Object Detection (HQOD) framework, which consists of two losses: Task-Correlated (TCorr) loss and Harmonious IoU (HIoU) loss. TCorr loss makes detectors focus more on optimizing samples with lower task harmony quality, and HIoU loss balances the optimization of the regression branch across different IoU levels during QAT. The combination of proposed losses can be conveniently integrated into various state-of-the-art QAT methods, effectively enhancing the task harmony of low-bit detectors and improving their overall performance. The future work can be shifted to other more complex task domains such as semantic segmentation, 3D object detection, and so forth. We think that similar phenomena of exacerbated task inharmony after low-bit quantization also exist in these domains.

\section*{Acknowledgment}
We thank anonymous reviewers for their kind advice on this research. This work is supported by the National Key Research and Development Program of China (2020AAA0109700), the National Science Fund for Distinguished Young Scholars (62125601), and the National Natural Science Foundation of China (62076024, 62006018).

\bibliographystyle{IEEEtran}
{\bibliography{IEEEexample}}

\begin{thebibliography}{10}
\providecommand{\url}[1]{#1}
\csname url@samestyle\endcsname
\providecommand{\newblock}{\relax}
\providecommand{\bibinfo}[2]{#2}
\providecommand{\BIBentrySTDinterwordspacing}{\spaceskip=0pt\relax}
\providecommand{\BIBentryALTinterwordstretchfactor}{4}
\providecommand{\BIBentryALTinterwordspacing}{\spaceskip=\fontdimen2\font plus
\BIBentryALTinterwordstretchfactor\fontdimen3\font minus \fontdimen4\font\relax}
\providecommand{\BIBforeignlanguage}[2]{{%
\expandafter\ifx\csname l@#1\endcsname\relax
\typeout{** WARNING: IEEEtran.bst: No hyphenation pattern has been}%
\typeout{** loaded for the language `#1'. Using the pattern for}%
\typeout{** the default language instead.}%
\else
\language=\csname l@#1\endcsname
\fi
#2}}
\providecommand{\BIBdecl}{\relax}
\BIBdecl

\bibitem{SSD}
W.~Liu, D.~Anguelov, D.~Erhan \emph{et~al.}, ``Ssd: Single shot multibox detector,'' in \emph{ECCV}, 2016.

\bibitem{RetinaNet}
T.-Y. Lin, P.~Goyal, R.~Girshick \emph{et~al.}, ``Focal loss for dense object detection,'' in \emph{ICCV}, 2017.

\bibitem{ATSS}
S.~Zhang, C.~Chi, Y.~Yao \emph{et~al.}, ``Bridging the gap between anchor-based and anchor-free detection via adaptive training sample selection,'' in \emph{CVPR}, 2020.

\bibitem{YOLOX}
Z.~Ge, S.~Liu, F.~Wang \emph{et~al.}, ``Yolox: Exceeding yolo series in 2021,'' \emph{arXiv:2107.08430}, 2021.

\bibitem{TOOD}
C.~Feng, Y.~Zhong, Y.~Gao \emph{et~al.}, ``Tood: Task-aligned one-stage object detection,'' in \emph{ICCV}, 2021.

\bibitem{TBD}
R.~Tang, Z.~yu~Liu, Y.~Li \emph{et~al.}, ``Task-balanced distillation for object detection,'' \emph{Pattern Recognit.}, 2022.

\bibitem{HT-CDOD}
J.~Deng, D.~Xu, W.~Li \emph{et~al.}, ``Harmonious teacher for cross-domain object detection,'' in \emph{CVPR}, 2023.

\bibitem{LSQ}
S.~K. Esser, J.~L. McKinstry, D.~Bablani \emph{et~al.}, ``Learned step size quantization,'' in \emph{ICLR}, 2020.

\bibitem{chun2023survey}
C.~Yang, R.~Zhang, L.~Huang \emph{et~al.}, ``A survey of quantization methods for deep neural networks,'' \emph{Chinese Journal of Engineering}, 2023.

\bibitem{DSQ}
R.~Gong, X.~Liu, S.~Jiang \emph{et~al.}, ``Differentiable soft quantization: Bridging full-precision and low-bit neural networks,'' in \emph{ICCV}, 2019.

\bibitem{TQT}
S.~Jain, A.~Gural, M.~Wu \emph{et~al.}, ``Trained quantization thresholds for accurate and efficient fixed-point inference of deep neural networks,'' in \emph{MLSys}, 2020.

\bibitem{AQD}
P.~Chen, J.~Liu, B.~Zhuang \emph{et~al.}, ``Aqd: Towards accurate quantized object detection,'' in \emph{CVPR}, 2021.

\bibitem{ResNet}
K.~He, X.~Zhang, S.~Ren \emph{et~al.}, ``Deep residual learning for image recognition,'' in \emph{CVPR}, 2016.

\bibitem{VOC}
M.~Everingham, S.~A. Eslami, L.~Van~Gool \emph{et~al.}, ``The pascal visual object classes challenge: A retrospective,'' \emph{IJCV}, 2015.

\bibitem{COCO}
T.-Y. Lin, M.~Maire, S.~Belongie \emph{et~al.}, ``Microsoft coco: Common objects in context,'' in \emph{ECCV}, 2014.

\bibitem{STE}
Y.~Bengio, N.~L{\'e}onard, and A.~Courville, ``Estimating or propagating gradients through stochastic neurons for conditional computation,'' \emph{arXiv:1308.3432}, 2013.

\bibitem{MobileNetv2}
A.~G. Howard, M.~Zhu, B.~Chen \emph{et~al.}, ``Mobilenets: Efficient convolutional neural networks for mobile vision applications,'' \emph{arXiv:1704.04861}, 2017.

\bibitem{CSPNet}
C.-Y. Wang, H.-Y.~M. Liao, Y.-H. Wu \emph{et~al.}, ``Cspnet: A new backbone that can enhance learning capability of cnn,'' in \emph{CVPR Workshop}, 2020.

\bibitem{Imagenet}
J.~Deng, W.~Dong, R.~Socher \emph{et~al.}, ``Imagenet: A large-scale hierarchical image database,'' in \emph{CVPR}, 2009.

\bibitem{MMDetection}
K.~Chen, J.~Wang, J.~Pang \emph{et~al.}, ``Mmdetection: Open mmlab detection toolbox and benchmark,'' \emph{arXiv:1906.07155}, 2019.

\bibitem{MQBench}
Y.~Li, M.~Shen, J.~Ma \emph{et~al.}, ``Mqbench: Towards reproducible and deployable model quantization benchmark,'' in \emph{NeurIPS}, 2021.

\bibitem{HarDet}
K.~Wang and L.~Zhang, ``Reconcile prediction consistency for balanced object detection,'' in \emph{ICCV}, 2021.

\end{thebibliography}

\end{document}